\documentclass[10pt,twocolumn,letterpaper]{article}

 \usepackage[pagenumbers]{cvpr} % To force page numbers, e.g. for an arXiv version
\newcommand{\Poincare}{Poincar\'e}
\newcommand{\Mobius}{M\"{o}bius}
\definecolor{cvprblue}{rgb}{0.21,0.49,0.74}
\usepackage[pagebackref,breaklinks,colorlinks,allcolors=cvprblue]{hyperref}
\usepackage{float}
\usepackage{makecell}
\usepackage{multirow}
\usepackage{tabularx}
\usepackage{placeins}
\usepackage{afterpage}

\def\paperID{*****} % *** Enter the Paper ID here
\def\confName{CVPR}
\def\confYear{2025}

\title{Deep Change Monitoring: A Hyperbolic Representative Learning Framework and a Dataset for Long-term Fine-grained Tree Change Detection}

\author{Yante Li\\
Univeristy of Oulu\\
{\tt\small yante.li@oulu.fi}
\and
Hanwen Qi\\
Wuhan University\\
{\tt\small hanwen.qi@whu.edu.cn}
\and
Haoyu Chen\\
Univeristy of Oulu\\
{\tt\small haoyu.chen@oulu.fi}\\
\and
Xinlian Liang\\
Wuhan Universityu\\
{\tt\small xinlian.liang@whu.edu.cn}
\and
Guoying Zhao\\
Univeristy of Oulu\\
{\tt\small guoying.zhao@oulu.fi}
}
\begin{document}
\maketitle

\begin{abstract}
%\vspace{-20pt}
In environmental protection, tree monitoring plays an essential role in maintaining and improving ecosystem health. However, precise monitoring is challenging because existing datasets fail to capture continuous fine-grained changes in trees due to low-resolution images and high acquisition costs. In this paper, we introduce UAVTC, a large-scale, long-term, high-resolution dataset collected using UAVs equipped with cameras, specifically designed to detect individual Tree Changes (TCs). UAVTC includes rich annotations and statistics based on biological knowledge, offering a fine-grained view for tree monitoring. 
To address environmental influences and effectively model the hierarchical diversity of physiological TCs, we propose a novel Hyperbolic Siamese Network (HSN) for TC detection, enabling compact and hierarchical representations of dynamic tree changes. 
 Extensive experiments show that HSN can effectively capture complex hierarchical changes and provide a robust solution for fine-grained TC detection. In addition, HSN generalizes well to cross-domain face anti-spoofing task, highlighting its broader significance in AI.  We believe our work, combining ecological insights and interdisciplinary expertise, will benefit the community by offering a new benchmark and innovative  AI technologies.   Source code is available on https://github.com/liyantett/Tree-Changes-Detection-with-Siamese-Hyperbolic-network.
\end{abstract}

\vspace{-20 pt}
\section{Introduction}
\vspace{-5 pt}
Forests, essential for climate stability and ecological health, act as crucial biological reserves \cite{eggleston20062006,woodwell1978biota}. Effective forestry and management hinge on diligent tree monitoring, which includes growth measurement, health assessment, and environmental impact analysis, and so on \cite{xiao2016individual}. One of the key aspects of tree monitoring is  Tree Change Detection (TCD). TCD identifies alterations in tree conditions, essential for tree management and overall forest ecosystem health.

\textit{This research aims to accomplish long-term fine-grained TCD with deep learning methods.}
Fine-grained TCD is critical yet challenging, especially when collecting long-term, high-resolution data in forests under diverse environmental conditions. Leveraging Unmanned Aerial Vehicles (UAVs) with cameras for forest imagery collection has become popular for its convenience and accessibility, offering an efficient alternative to traditional methods like LiDAR for large-scale data collection, essential for monitoring trees and ecological sustainability \cite{tang2015drone}. Thus, to boost data-driven learning methods for the research field, this paper introduces a UAV-camera-based dataset tailored for detecting individual Tree Changes (UAVTC), enabling precise tree monitoring, as shown in Figure \ref{fig:Demonstration} (a).  \textit{To our knowledge, UAVTC is the first dataset to provide detailed insights into precise TCs over time, supporting sustainable forest management and ecological research.}

\begin{table*}[!ht]
	\centering
	\caption{\textcolor{black}{ UAV-camera-based tree datasets. Our dataset offers large-scale, high-resolution tree images accompanied by fine-grained TC annotations. In terms of resolution ($cm/pixel$),  the lower value represents the better performance.}} %\hspace{\textwidth} [w$\backslash$o MM means without motion magnification, while w$\backslash$ MM means with motion magnification].}
 \vspace{-5pt}
  \scalebox{0.92}{
	\begin{tabular}{|c|c |c| c|  c| c|  c| c|c|c| }
		\hline\
		Dataset & Task& \makecell*[c]{Resolution $\downarrow$ \\($cm/pixel$) } &\makecell*[c]{Test site\\size ($m$)}  & \makecell*[c]{Duration\\(Month)}  & \makecell*[c]{Flight\\days} & \makecell*[c]{Image /\\Image pairs}&\makecell*[c]{Publicly\\available}    \\ \hline
        \cite{park2019quantifying}&Leaf coverage  &7&1000×500&12&34& 2,422&$\times$\\ \hline
        \cite{araujo2021strong}&Treefall detection&5&1000×500&60&60& 60 &\checkmark  \\ \hline
        \cite{lee2023cost}&Flower recognition&5 &1000×500&24&21& 34,713 &$\times$ \\ \hline
        UAVTC (Ours) &TCD (color, blossom, leaf, branch) &0.5 &110×140&12&85& 245,616&\checkmark  \\ \hline
	\end{tabular}}

  {\raggedright{}  $\checkmark$ indicates the datasets is publicly available; $\times$ signifies it's not.\par
  \raggedright {\,}
   \vspace{-10pt}
  {$\downarrow$} {represents the lower value with the better performance.} \par
  }
 \label{dataset_compare}
 \vspace{-5pt}
\end{table*}

%Efficient fine-grained tree change detection is crucial, yet it faces challenges due to the (1) difficulty in high-resolution tree data collection, (2) variable weather conditions, and (3) the intrinsic complexity of tree dynamics.

\textit{The primary challenge in TCD lies in identifying  extrinsic changes driven by the environments and intrinsic changes referring to  physiological processes.} As shown in Figure \ref{fig:Demonstration} (c),  environmental factors such as fog and direct sunlight can obscure views and cast shadows, leading to non-tree-related changes. These extrinsic changes can mislead and result in incorrect TCD. Apart from extrinsic changes, physiological TCs are inherently hierarchical, involving both growth and decay processes.  These changes can be classified into four main categories: changes in color, leaf transformations, flowering and fruiting cycles, and damage caused by humans, as shown in Figure \ref{fig:Demonstration} (b). These diverse attributes of trees give rise to complex hierarchical relationships between TCs caused by environmental and physiological factors, which are not effectively captured by Euclidean geometry. Instead, hyperbolic geometry provides a solution by faithfully representing tree-like structures \cite{khrulkov2020hyperbolic}, making it ideal for modeling and understanding physiological TCs.

To address the above issues, we propose a Hyperbolic Siamese Network (HSN) designed to effectively model hierarchical changes and mitigate the impact of environmental factors, thereby enabling efficient identification of physiological TCs. HSN achieves this by (1) \textbf{Modeling Hierarchical Structure}: HSN encodes hierarchical relationships present in changes driven by environmental factors and intrinsic physiological changes at distinct levels, clarifying the pathways and impacts of each change; (2) \textbf{Optimizing Data Embedding and Representation}: Hyperbolic space allows for an efficient low-dimensional representation of high-dimensional hierarchical relationships, enabling both environmental and physiological changes to be effectively embedded and analyzed for structural and influential similarities or differences; (3) \textbf{Modeling Spatiotemporal Dynamics}: To model changes between tree images taken at different time points, we designed a network that integrates a Siamese structure within hyperbolic space. The Siamese network structure within hyperbolic space enhances analysis of dynamic changes across time and space, enabling clearer insights into trees' responses to complex interactions between environmental conditions and physiological factors.

To the best of our knowledge, this is the first hyperbolic network designed specifically for change detection. \textit{Thus, we also provide extra experiments on cross-domain face anti-spoofing (CD-FAS), showing that HSN excels at managing complex real-world changes involving extrinsic and intrinsic patterns.}

The main contributions are as follows: 
\begin{itemize}
    \item We collect a large-scale UAVTC dataset for long-term fine-grained TCD. To the our best knowledge, the UAVTC is the first TCD dataset that is collected from UAV devices, which contains rich annotations and statistics, covering multiple aspects of tree growth and health, enabling fine-grained tree monitoring over time.
    \item  We propose a novel Hyperbolic Siamese Network to model tree changes in hyperbolic space, overcoming misleading caused by extrinsic factors and effectively capturing the intrinsic tree changes. 
    \item Extensive experiments demonstrate that HSN is highly effective for fine-grained tree monitoring and can be generalized to CD-FAS task, highlighting its versatility and robustness.
\end{itemize}

%To further demonstrate the effectiveness and impact of HSN, we evaluate its performance on cross-domain face anti-spoofing tasks, highlighting its significance in AI.

  \vspace{-10pt}
\section{Related work}
\subsection{Tree monitoring dataset comparison }
Forest monitoring is crucial for effective tree management and policy development. Remote sensing provides continuous, all-weather, multi-spectral observations to detect forest changes with precision. While spaceborne platforms offer broad land cover data, their limited resolution is insufficient for detailed forest analysis \cite{friedl2002global,shao2006satellite}. In contrast, UAVs provide more frequent and closer-range observations \cite{ambrosia2011unmanned,shahbazi2014recent}.

UAVs equipped with LiDAR technology are capable of generating accurate 3D models of trees, offering key data such as tree height, area, and volume growth through spatial coordinates and surface reflectivity measurements \cite{tang2023monitoring,xiao2016individual}. However, LiDAR lacks the capability to capture color or intricate texture details, thus limiting its ability to reflect TCs \cite{richardson2011strengths}.

Instead, UAVs equipped with RGB cameras excel in capturing long-term tree changes that include color and texture details. Park et al. \cite{park2019quantifying} conducted leaf phenology research in a 50 ha tropical forest using a UAV equipped with an RGB camera. Similarly, Lee et al. \cite{lee2023cost}  employed this technology to identify flowering patterns of various tree species over five years with 34,713 images.  Additionally, a five-year dataset of monthly drone-captured RGB images was analyzed to track treefall and branchfall in Panama \cite{araujo2021strong}, offering deep insights into forest dynamics. 
However, the above studies focus on singular aspect of tree states, such as blossoming or branch falling, without a comprehensive view of the TCs over time. Comparing trees over time is vital for understanding growth, health, and environmental responses, enabling early disease detection and guiding forest management, ecological research, and conservation efforts, ultimately crucial for biodiversity and ecosystem sustainability.

To bridge this gap, our paper presents the UAVTC dataset, facilitating the observation of fine-grained TCs across multiple aspects and allowing for the comparison of individual trees over time.
The UAVTC includes color changes, branch modifications, blossom, and more, averaging nearly once a week with a resolution of 0.5 cm/pixel. It presents a unique resource for achieving detailed tree monitoring with 245,616 tree image pairs, covering a wider spectrum of TCs than previously available datasets. The specific comparison of UAVTC with other existing datasets is illustrated in Table \ref{dataset_compare}.

\subsection{Hyperbolic geometry}

Hyperbolic geometry, characterized by a constant negative Gaussian curvature, has proven to be beneficial for hierarchical relationships between images, and representing tree-like structures, taxonomies, and graphs \cite{khrulkov2020hyperbolic,nickel2017poincare,aly2019every}. There are five isometric models of hyperbolic geometry \cite{cannon1997hyperbolic}: the  Lorentz (hyperboloid) model, Klein model, Hemisphere model, {\Poincare}  ball,  and {\Poincare}  ball half-space model. Among these models,  the {\Poincare}  ball model has been dominated in the hyperbolic deep neural networks due to its differentiable distance function and the simplicity of its constraints on representations.

%the Lorentz ball model and the {\Poincare}  ball model are the most commonly used models in machine learning \cite{nickel2017poincare}. 

Maximillian et al. \cite{nickel2017poincare} first introduced the {\Poincare}  model for embedding learning, taking into account latent hierarchical structures.  Their research demonstrated that {\Poincare}   embeddings exhibit superior performance compared to Euclidean embeddings, especially in datasets characterized by latent hierarchies. Recently, hyperbolic embeddings have 
been very successful in the natural language processing field 
\cite{nickel2018learning,sarkar2011low,sala2018representation}.  However, the above works utilized Riemannian optimization algorithms to embed individual words into hyperbolic space due to the discrete nature of data in natural language processing. Extending this method to visual data proves challenging, as image representations are commonly generated with convolutional neural networks \cite{khrulkov2020hyperbolic}.

Khrulkov et al. \cite{khrulkov2020hyperbolic} investigated the hybrid architecture with the convolutional neural networks operating in Euclidean space and only the final layers operating in hyperbolic space. This research verifies the effectiveness of capturing semantic and hierarchical information in images by the hyperbolic network. Then, Hyperbolic Networks (HNs) have been successfully developed for various computer vision tasks including image segmentation \cite{atigh2022hyperbolic,chen2023hyperbolic}, action recognition \cite{peng2020mix}, face recognition \cite{trpin2022face,han2023hyperbolic}, and metric learning \cite{ermolov2022hyperbolic,guo2021free,ge2023hyperbolic}. In summary, hyperbolic geometry proves instrumental in diverse applications, and its incorporation into deep learning frameworks enhances the representation of hierarchical relationships and semantic similarities in various types of data, including images, text, and graphs. Considering the complex hierarchical relationships in intrinsic TCs and noises caused by environments, this paper presents utilizing hyperbolic space to effectively analyze and understand TCs. The HSN offers a new perspective and enhanced capabilities for capturing the complex and hierarchical changes.

%They also proved that Poincare embeddings can outperform Euclidean embeddings significantly on data with latent hierarchies, both in terms of representation capacity and in terms of generalization ability. 

 %Initially, representations are learned in the hyperbolic space, and subsequently, hyperbolic geometry is integrated into deep learning through hyperbolic neural networks. Recent developments in hyperbolic image embeddings involve the addition of shallow hyperbolic layers to Euclidean deep convolutional neural networks.

%Hyperbolic deep neural networks, introduced by Nickel and Kiela (2017) and further developed by Ganea et al (2018), provide powerful geometric representations with neural structures constructed within the space of hyperbolic geometry. The approach involves learning representations in hyperbolic space, thereby integrating hyperbolic geometry into deep learning.

\section{UAV-camera-based TC dataset collection}

\subsection{\textcolor{black}{Test site and equipment}}
To study fine-grained TCs, this paper presents a comprehensive tree dataset by UAV equipped with an RGB camera.
The experiment was carried out in a $110 m \times 140 m$ plot in a mixed urban forest. The species mainly include Camphora officinarum, Cedrus, firethorn, and Malus mandshurica. Some species are evergreen and non-flowering plants, such as Camphora officinarum and Cedrus, while other species may blossom and even yield fruit. 

\begin{figure}
\centerline{\includegraphics[width=8.5cm]{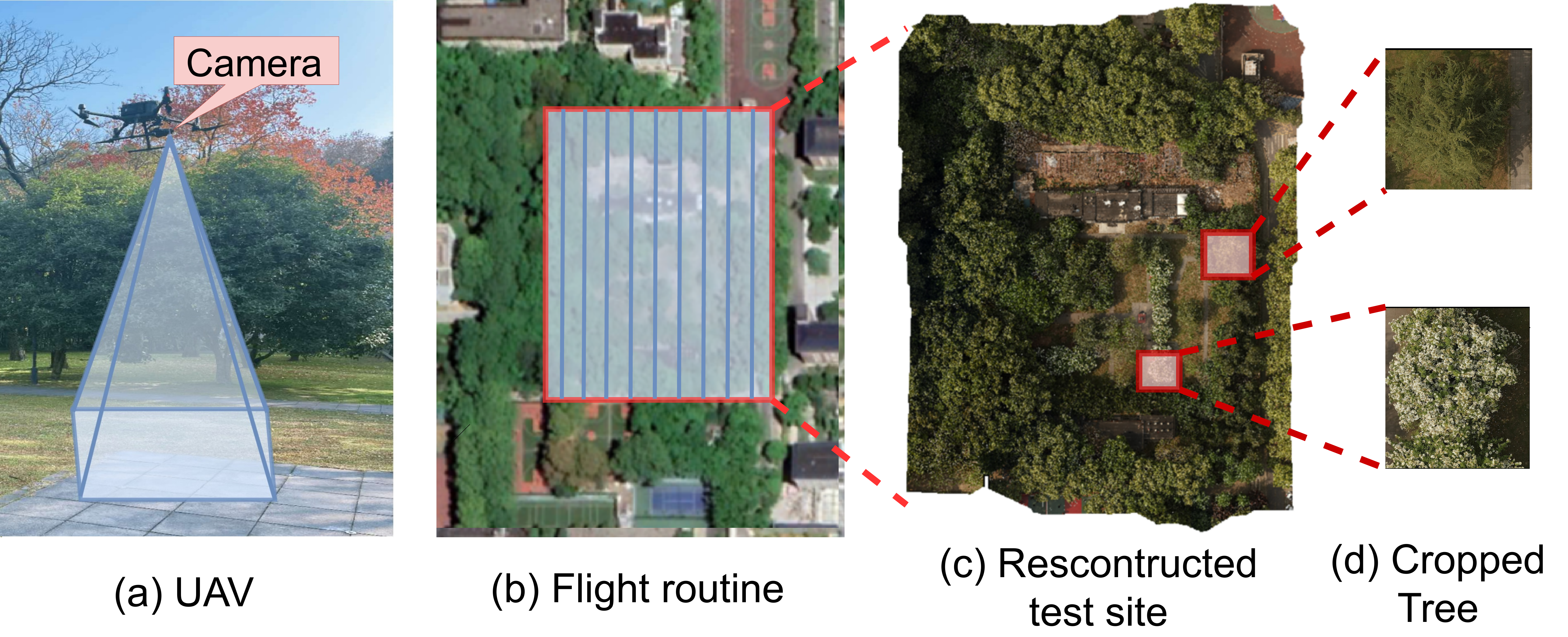}}
\vspace{-10pt}
\caption{The illustration of the data collection process. (a) the UAV equipped with a camera (DJI Zenmuse P1), (b) the flight routine outlining the test site, (c) the reconstructed test site after the UAV has captured the necessary imagery, and (d) the final cropped tree images. }   
\label{fig:dataset_collection}
\vspace{-10pt}
\end{figure}

The platform utilized in this paper is DJI M300 RTK (DJI Innovations), and the camera is DJI Zenmuse P1 (lens FOV 63.5 degrees, focus length $35 mm$, Max image size $8192 \times 5460$ pixels), As shown in Figure \ref{fig:dataset_collection} (a). Both the front and side overlapping in between photos were set to $80\%$ and the flight altitude was $50 m$ above the ground, and the flight speed was $3m/s$. The shutter was set to $1/400$ on bright days, when the sunlight was not sufficient, the shutter could be set to $1/200$. In addition, ISO was set to $100$ and F was $2.8$. 
During the data collection process, the UAV followed a predefined flight routine, as illustrated in Figure \ref{fig:dataset_collection} (b). Typically, each flight mission resulted in the capture of approximately 220 pictures of the test site. 
Moreover, Digital Orthophoto Model (DOM) was produced by DJI Terra (Figure \ref{fig:dataset_collection} (c)), and geographic registration root mean square error was within $4 cm$, RMS of reprojection error was within $1 px$. For more details about the dataset collection, please refer to the supplementary material.

\subsection{\textcolor{black}{Preprocessing and annotation}}
\vspace{-5pt}
The goal is to detect changes in individual trees, necessitating the acquisition of their respective regions of interest (ROIs). Each tree was individually outlined with a rectangle in ENVI software, based on GPS coordinates from the tree data gathered on 07/11/2022. Since each DOM was projected to the same coordinate system, it was straightforward to crop according to the ROI. The ROI was determined using the four corners of a rectangle, slightly larger than the tree crown to ensure full coverage, accounting for potential crown growth.
Figure \ref{fig:dataset_collection} (c) illustrates two representative cropping of trees, the red transparent rectangles are the ROIs, and the two individual trees are shown in Figure \ref{fig:dataset_collection} (d).

To evaluate the change, the annotation process is divided into two steps. Firstly, two forest experts independently annotated the tree's states.  In cases of disagreement, the final state was determined through a consensus reached by these experts after discussion. Specifically, a total of six states are considered including green leaf, yellow leaf, branch, fallen leaves or sprout, blossom, and destroy. The inconsistent states indicate the presence of tree changes, otherwise is no change. For more details about the data annotation, please refer to the supplementary material.

%Figure X illustrate the six labels. Specifically, fallen leaves and sprout share the same label, it is because that these two statuses represent interim period, indicating the tree starts to make change.

%\vspace{-10pt}
\subsection{{Data statistics}}

The tree images were collected from 07/10/2022 to 03/11/2023. Data collection is usually performed between 2 p.m. and 4 p.m., as often as possible once every two days, but sometimes once every three or four days due to weather conditions, such as rain or strong winds, and on average two or three times a week. A total of 85 time points capturing the changes in trees over one year were collected.  68 trees were selected and cropped for analysis. The trees are in diversity, from evergreen to deciduous, flow-
ering, and fruit-bearing trees. Despite the limited number of
trees, which may influence the generalizability of the model,
the dataset’s diversity sufficiently supports robust
model training for individual tree change detection.
To thoroughly investigate tree changes, we perform a pairwise comparative analysis of tree images at all time points for each tree.
In total, we have 245,616 pair tree images. $72\%$ belongs to change and $28\%$ belongs to no change. 
The average pixels for each tree ROI is around $2000 \times 2000$ pixels.

\begin{figure}
\centerline{\includegraphics[width=8.5cm]{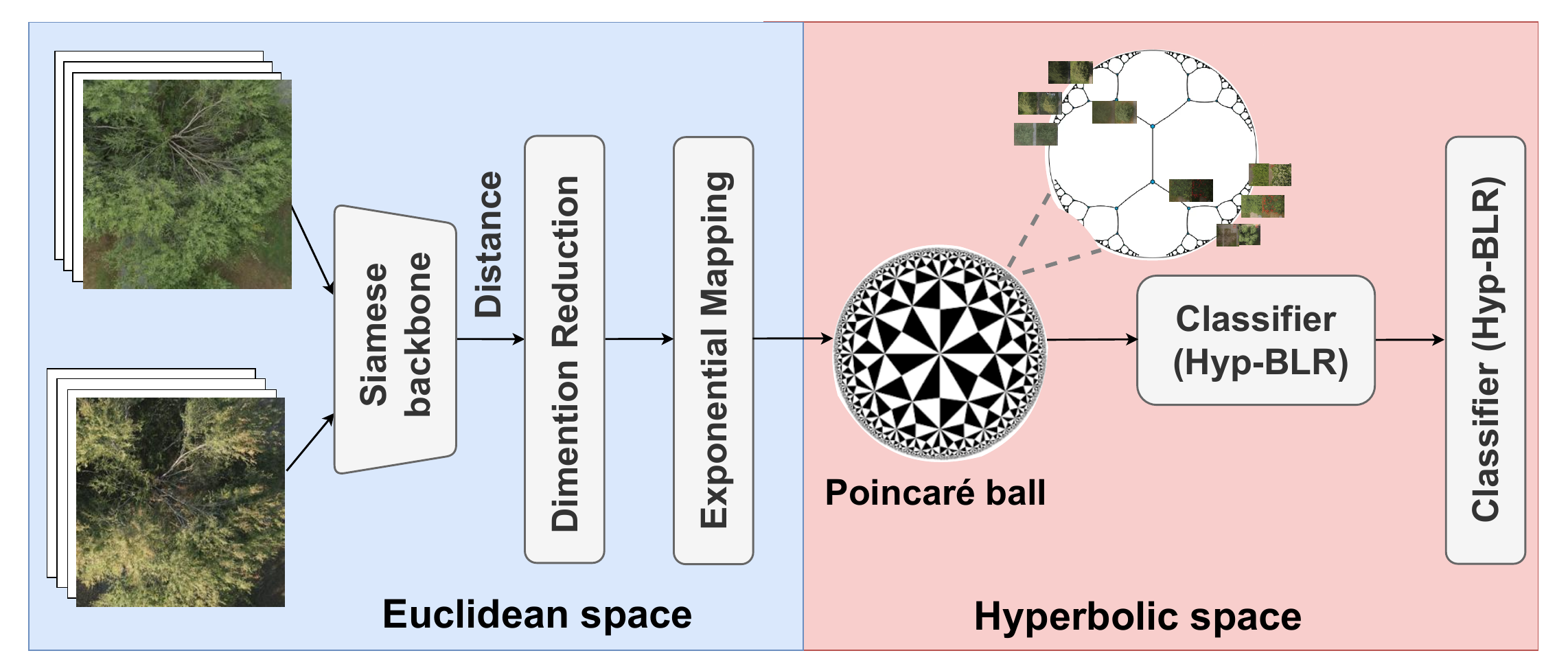}}
\caption{Framework of Hyperbolic Siamese Network. First, features extracted from the backbone Siamese network undergo a comparison process to compute change. Subsequently, a fully connected (FC) layer reduces the dimension of the change feature. The model utilizes exponential mapping to transform embeddings from the Euclidean space to hyperbolic space. 
A Hyp-BLR is followed for the classification
Finally, a Hyp-BCE loss function is employed to train modes for TCD.  }
\label{fig:Framework}
\vspace{-10pt}
\end{figure}

\section{Methodology}

% The examples on {\Poincare}  ball with blue rectangles represent trees without change, while the green ones present trees with change.

In this section, we introduce the technical details of the HSN for TCD. The schematic overview of the proposed method is depicted
in Figure \ref{fig:Framework}. First, we introduce the operations in hyperbolic spaces in Section \ref{sec:hyperbolic}. Hyperbolic feature clipping for model optimization is described in Section \ref{sec:clipping}. Then we describe
the hyperbolic Siamese architecture.
  Hyperbolic cross-entropy loss is introduced in Section \ref{sec:loss}. 
Finally, we compute the $\delta$-hyperbolicity on the embeddings of TCs, theoretically verifying the presence of hierarchical structures within the TCs.

\subsection{Hyperbolic operations}
\label{sec:hyperbolic}
As highlighted in the Introduction section, TCD is significantly influenced by environment conditions and characterized by hierarchical structures. Prior studies have shown that hyperbolic geometry effectively captures data hierarchies, enhancing robustness and accuracy by mitigating the impacts of extrinsic factors like weather variants\cite{khrulkov2020hyperbolic}. Therefore, hyperbolic geometry is utilized in this paper.  Specifically, our research utilizes the {\Poincare} ball model \cite{nickel2017poincare}, as the {\Poincare}  ball possesses a differentiable distance function and is characterized by a relatively simple constraint on its representations. These strengths make it suitable for TCD research.

%Hyperbolic space, also known as negatively curved space, is defined by a Riemannian manifold with a constant negative curvature.

Formally, the n-dimensional  {\Poincare}  ball $\left(\mathbb{H}_c^{n},g^{\mathbb{H}} \right)$ is defined by the manifold $\mathbb{H}_c^{n}$ as

\begin{equation}
\mathbb{H}_c^{n} =\left \{ X\in \mathbb{R}^{n}:c\left \| x \right \|^{2}< 1, c\ge 0    \right \},
\end{equation}
and the Riemannian metric $g^{\mathbb{H}}$ is 
\begin{equation}
g_x^{\mathbb{H}}  = {\lambda _x^{c} }^{2} g^{\mathbb{E} },
\end{equation}
where $g^{\mathbb{E}}$ is the  Euclidean metric tensor and ${\lambda _x^{c} }$ is the conformal factor defined as  ${\lambda _x^{c}= \frac{2}{1-c\left \| x \right \| ^{2} } }$.  $g^{\mathbb{H}}$  is associated with tangent space 
$T_x$.
Note, $c^{-\frac{1}{2} }$  is the radius of {\Poincare}  ball.  $c$ controls the the curvature of the {\Poincare} ball. As  $c$ approaches 0, operations revert to Euclidean geometry.

Hyperbolic spaces differ from traditional vector spaces, making standard operations like addition and multiplication unusable. {\Mobius} gyrovector spaces offer a solution to generalize these operations. Key operations for hyperbolic networks are described below.

\vspace{5pt}
%\subsubsection{{\Mobius} addition}
\noindent \textbf{{ {\Mobius} addition.}} {\Mobius} gyrovector spaces offer a means to define operations like addition and multiplication. These operations are fundamental in hyperbolic networks \cite{khrulkov2020hyperbolic}.
For a pair $x, y \in \mathbb{H}_{c}^{n}$,
their addition is defined as
\begin{equation}
x\oplus_{c}y =\frac{(1+2c \langle x, y\rangle +c\left\|  y\right\|^{2})x +(1-c\left\| x\right\|^{2})y}{1+2c\langle x,y\rangle +c^{2}\left\|  x\right\|^{2} \left\| y \right\|^{2}}.
\end{equation}

\noindent  \textbf{{Distance.}}
In hyperbolic geometry, the distance between two points, $x, y \in \mathbb{H}_{c}^{n}$, is defined as
\begin{equation}
d_{hpy}(x,y) =\frac{2}{\sqrt{c}}arctanh(\sqrt{c}\left\|  -x\oplus_{c}y\right\|).
\end{equation}
When c approaches 0, the hyperbolic distance function converges to the Euclidean distance. This convergence is expressed as $\lim_{c \to 0} =d_{hpy}(x,y) =2\left \| x-y \right \|$.

\vspace{5pt}

\noindent  \textbf{Exponential map.}
To conduct operations in hyperbolic space, it is necessary to establish a bijective mapping between Euclidean vectors and hyperbolic vectors in the {\Poincare}  ball model. The pivotal tool for this purpose is the exponential map. The map establishes a one-to-one correspondence between points from Euclidean space to points in hyperbolic space. The exponential map is formulated as:
\begin{equation}
exp_{x}^{c}(v)  =x\oplus_{c}(tanh(\sqrt{c}\frac{\lambda^{c}_{x}\left\| v \right\|}{2})\frac{v}{\sqrt{c}\left\| v \right\|}). 
\label{equ:mapping}
\end{equation}

\noindent  \textbf{Hyperbolic binary logistic regression (Hpy-BLR).}
In the context of the TCD task, two classes are established: when $k=0$, it indicates the absence of change, and when 
$k=1$, it denotes the presence of change and  $p_k \in \mathbb{H}_{c}^{n}$, $t_k \in T_{p_k} \mathbb{H}_{c}^{n} \setminus\{0\}$.  The formulation of binary logistic regression within the {\Poincare}  ball is expressed as follows:

\vspace{5pt}
\begin{equation}
p(y=k\mid x)\propto exp(\frac{\lambda_{pk}^{c}\left \| t_{k}  \right \| }{\sqrt{c} } arcsinh ( \frac{2\sqrt{c}\left \langle -p_k\oplus_c x,t_k  \right \rangle  }{(1-c\left \| -p_k\oplus_c x \right \|^{2})\left \| t_k \right \| } )).
\label{equ:logit}
\end{equation}

\noindent  \textbf{Feature clipping.}
\label{sec:clipping}
Previous research \cite{guo2021free,guo2022clipped} provides empirical evidence indicating that HNs often experience the issue of vanishing gradients. This occurs because HNs tend to drive embeddings towards the boundary of the {\Poincare} ball, resulting in the gradients of Euclidean parameters becoming extremely small. To mitigate potential numerical errors, a fixed threshold is applied to the norm (magnitude) of the vectors to prevent them from reaching extreme values \cite{guo2021free}, following the equation below:

\vspace{-6pt}
\begin{equation}
C(x^{E}; r)  =min\left \{ 1,\frac{r}{\left \| x^{E}  \right \| }  \right \} \cdot x^{E}
\end{equation}

where $x^{E}$ lies in the Euclidean space and $x_{C}^{E}$ represents its clipped counterpart. The 
hyperparameter $r$ denotes a novel effective radius within the {\Poincare}  ball. This feature clipping method imposes a hard constraint on the maximum norm of the hyperbolic embedding which avoids the inverse of the Riemannian metric tensor approaching zero \cite{guo2021free}.

\subsection{Hyperbolic Siamese framework}
\label{sec:framework}

What is the desired representation for TC? We argue that such representation should
effectively present the precise physiological change in trees and maintain robustness to  changes caused by extrinsic factors. 
To achieve the above goals, we train an HSN (shown in Figure \ref{fig:Framework}) consisting of Siamese framework and hyperbolic geometry. The key objectives of this approach are 1) accurately evaluating TCs and 2) learning robust features avoiding misleading influences from extrinsic factors.

%To achieve fine-grained TCD, we need to compare trees at different times. The Siamese network is a powerful approach that allows for the direct comparison of pairs of data samples, making it particularly well-suited for tasks like TCD \cite{chopra2005learning,lecun2005loss}.  However, the training of Siamese network requies and easily influenced 

For a classical Siamese network \cite{chopra2005learning,lecun2005loss},  the input is a pair of images, represented as $I_1$ and $I_2$.  The Siamese network learns to project inputs into a space where the distance between similar inputs is minimized, and the distance between dissimilar inputs is maximized, facilitating tasks such as similarity comparison or verification  \cite{chen2021exploring}. Specifically, $f(I;W)$ is the output of the Siamese network, where $W$ represents the parameters of the network. A common choice the measure the feature distance is the Euclidean distance:
\vspace{-5pt}
\begin{equation}
\vspace{-2pt}
D(I_1,I_2) = \left \| f(I_1;W)-f(I_2;W) \right \|,
\end{equation}

However, as discussed in the introduction section, TCs include a hierarchical structure that can not be effectively estimated by the flat Euclidean space. Instead, hyperbolic space characterized by negative curvature has been found to be well-suited for representing hierarchical relationships. The hyperbolic space is leveraged to represent complex TCs with hierarchy. 
Firstly, a fully connected (FC) layer is employed to reduce the dimension of the change feature. Then the Euclidean distance embedding is transformed to hyperbolic distance by exponential mapping with Equation \ref{equ:mapping}. A Hyp-BLR layer is followed for classification. Finally, the Hyp-BCE loss function (Equation (\ref{equ:loss})) is employed to identify and assess the changes in trees.

%The Siamese network is trained by defining a loss function that penalizes large distances for similar inputs and small distances for dissimilar inputs.

\subsection{Hyperbolic binary cross
entropy loss }
\label{sec:loss}
The Hyperbolic Binary Cross
Entropy (Hpy-BCE) loss is calculated using the predicted probability $\hat{p_i}$ obtained from Equ. \ref{equ:logit}. The loss is defined as:
\vspace{-5pt}
\begin{equation}
\vspace{-2pt}
{L_a} ={p_i}log(\hat{p_i})+(1-p_i)log(1-\hat{p_i}), \\
\label{equ:loss}
\end{equation}
where $p_i$ is the ground truth for changes in a tree, with 1 denoting a change in the tree and 0 denoting no change. $\hat{p_i}$ is the predicted probability of TC.

\subsection{ $\delta$-Hyperbolicity }
\label{sec:delta}
\vspace{-2pt}
In this section, we validate the presence of hierarchical structures in TC data through $\delta$-hyperbolicity \cite{khrulkov2020hyperbolic}. This metric serves to quantify the similarity in data structure between Euclidean and hyperbolic space. The $\delta$-hyperbolicity values range from 0 to 1, where a calculated value closer to 0 indicates a high degree of hyperbolicity in the data, signifying a strong hierarchy. Conversely, a computed $\delta$-hyperbolicity closer to 1 suggests the absence of hierarchy in the dataset. 

This hyperbolicity evaluation is made through
Gromov product \cite{fournier2015computing}:
\vspace{-10pt}

\begin{equation}
\vspace{-2pt}
(y, z)_x =\frac{1}{2} (d(x, y) + d(x, z)-d(y, z))
\label{equ:gromov}
\end{equation}
where  $x, y, z \in \chi $ and $\chi$ is the arbitrary (metric) space endowed with
the distance function $d$. Given a set of
points, we compute the pairwise Gromov products and represent them in the form of a matrix $A$. Subsequently, the value of $\delta$ is identified as the maximum element in the matrix obtained from the min-max matrix product operation, denoted as $(A\oplus A)-A$. In this context, the symbol $\oplus$ represents the min-max matrix product, which is formally defined as $(A \oplus B)_{ij} = max_kmin \left \{A_{ik}, B_{kj}\right \} $ \cite{fournier2015computing}.
%The value of $\delta$ is related to the optimal radius of the Poincare ball concerning embeddings by the formula: 
%\begin{equation}
%c(x)=(\frac{0.144}{\delta_X}) ^{2} 
%\label{equ:c}
%\end{equation}

We evaluate $\delta$ for tree image embeddings extracted by Siamese networks with ResNet18 \cite{he2016deep}, ResNet34, ResNet101, and VGG16 \cite{simonyan2014very}. Table \ref{table_delt} demonstrates the obtained relative $\delta$ values. Table \ref{table_delt} shows that all networks exhibit low $\delta$ (closer to 0 than 1), indicating high hyperbolicity in the tree change dataset. This observation suggests that TCD tasks can benefit from hyperbolic representations of images and the experiments in Section \ref{sec:guidelines} verify the effectiveness of hyperbolic representations.

\begin{table}[h]
	\centering
	\caption{$\delta$-Hyperbolicity. Low $\delta$ (closer to 0 than 1)  indicates a high degree of hyperbolicity in the embeddings. } %\hspace{\textwidth} [w$\backslash$o MM means without motion magnification, while w$\backslash$ MM means with motion magnification].} &Mobile

\scalebox{0.80}{{
	\begin{tabular}{|c|c |c| c| c| c| c|c|} 
		\hline\
	Network	& ResNet18 & ResNet34& ResNet101 &VGG16 \\ \hline
 $\delta$  &  0.172    & 0.144 &0.114 &0.212\\ \hline
           \end{tabular}}}
		\label{table_delt}
  \vspace{-15pt}
\end{table}

%Inception&SE-Mobilenet
%&0.206&0.208

\begin{table*}[h]
	\centering
	\caption{The impacts of the parameter $c$ used in HSNs.} %\hspace{\textwidth} [w$\backslash$o MM means without motion magnification, while w$\backslash$ MM means with motion magnification].}
 \vspace{-5pt}
 \scalebox{0.85}{
	\begin{tabular}{|c|c| c| c |c| c| c|c|c|c|c|c|c|c|c|} 
		\hline\
		\multirow{2}{*}{$c$}& \multicolumn{5}{c|}{ACCURACY} & \multicolumn{5}{c|}{F1}  \\ \cline{2-11}
	   & 0.1&0.3& 0.5&0.7& 1.0&0.1&0.3& 0.5&0.7& 1.0\\ \hline
      HS-ResNet18&  80.71& \textbf{80.80}&79.69 &79.74&80.27& \textbf{0.6156}&0.6053&0.5907&0.5754&0.6057           \\ \hline
	 HS-ResNet34&81.43&\textbf{81.86}&80.08&80.02&81.00 & \textbf{0.6479}&0.6449&0.5790&0.5813&0.6300         \\ \hline
      HS-ResNet101& {86.50}&85.60 & 86.18&86.15& \textbf{86.83}&0.7204&0.6877&0.7137&0.7127 & \textbf{0.7372}             \\ \hline
      HS-VGG16& 74.02& \textbf{77.18}&74.60&74.55&72.29&0.5805& \textbf{0.6101}&0.5841&0.5756&0.5455            \\ \hline
      HS-InceptionV3  &76.55&77.08& \textbf{77.23}&76.50&77.14
&0.6552&\textbf{0.6556}&0.6481&0.6447&0.6467           \\ \hline
      HS-MobileNet&77.05&76.73&76.79&77.56& \textbf{77.74}&0.4822&0.4979& \textbf{0.5048}&0.4858&0.4953     \\ \hline
 \end{tabular}}
 \label{table_hyc}
 \vspace{-10pt}
\end{table*}

\section{Experiments}
\label{sec:guidelines}
In this section, we first introduce the protocols, evaluation metrics, and implementation details in this paper.  Then, we evaluate the impacts of hyperparameters. Finally, We show the baselines on our constructed tree change dataset and the improvements with hyperbolic spaces on both TC and CD-FAS tasks.

\subsection{Protocols and evaluation matrix}
TCD task is to detect whether there are changes between trees at different time points to monitor climate change and protect the environment automatically. In the experiment, around two-thirds of the trees (44 trees) were used as the training data, while one-third (24 trees) were reserved for testing.  Since 85 flight days were collected, there are 158,928 image pairs for training and 86,688 image pairs for testing. 

Both accuracy and F1-score are utilized to evaluate the performance of TCD. In binary classification tasks, especially with imbalanced samples, it is better to incorporate F1-score with accuracy to interpret the
algorithm performance.

\subsection{Implementation details}
For training Euclidean Siamese networks (ESNs), we employ SGD with momentum, while HSNs are trained with the AdamW optimizer \cite{loshchilov2017decoupled}. 
All the networks are fine-tuned on ImageNet \cite{deng2009imagenet} and trained for 30 epochs. We use an initial learning rate of $1 \times 10^{-6}$
 for convolutional layers, $0.001$ for FC layers, and $0.1$ for hyperbolic layers. The learning rate decays by 10 every 10 epochs. The batch size is 128. In the training stage, the images are resized to $256 \times 256$  ($310 \times 310$ for HS-InceptionV3) and randomly cropped to $224 \times 224$ ($299 \times 299$ for HS-InceptionV3). In the testing stage, images are captured with a resolution of  $224  \times 224$ using a center crop ($299 \times 299$ for HS-InceptionV3). 
The clipping radius (defined in Section  \ref{sec:delta}) $r = 2.3$, following \cite{ermolov2022hyperbolic}. The random seed is set to 42.

%All the experiments

\subsection{Abalation study}
\label{sec:parameter}
%\subsubsection{Manifold curvature}
\textbf{Manifold curvature.} As illustrated in Sections \ref{sec:hyperbolic}, the value of $c$ represents the manifold curvature of a {\Poincare}  Ball model. To comprehensively analyze the impacts of the parameter $c$,  experiments were conducted on UAVTC dataset with  $c=\{0.1,0.3,0.5,0.7,1.0\}$. The feature dimensions were fixed with eight. 
Table \ref{table_hyc} shows the HSN performances depending on the curvature value $c$. From Table \ref{table_hyc}, we find that
the methods are relatively robust in the range $(0.1, 1.0)$. In general, the deeper networks could achieve better performance with higher $c$. Specifically, HS-ResNet101 performs best with $c=1.0$, while HS-ResNet18 and ResNet34 achieve optimal F1-scores with $c=0.1$. The results suggest that shallow networks tend to learn embeddings in a manifold with flatter curvature, whereas deep networks display a trend towards learning in manifolds with more pronounced curvature.

%In Sections \ref{sec:hyperbolic}, c represents the manifold curvature in a Poincaré Ball model. Experimenting with $c={0.1,0.3,0.5,0.7,1.0}$ on the Tree change dataset (Table \ref{table_hyc}), HS-networks exhibit robustness in the (0.1, 1.0) range. Deeper networks generally perform better with higher c. ResNet101 excels at c=1.0, while ResNet18 and ResNet34 achieve optimal F1-scores at c=0.1. Results suggest that tree change image embeddings contain hierarchical information, and shallow networks learn in flatter manifold curvature.
%Moreover, in section \ref{sec:delta}, the c has been roughly measured by Equ.\ref{equ:c}, as shown in Table \ref{table_delt}

%\vspace{5pt}

\begin{figure}
\centerline{\includegraphics[width=8cm]{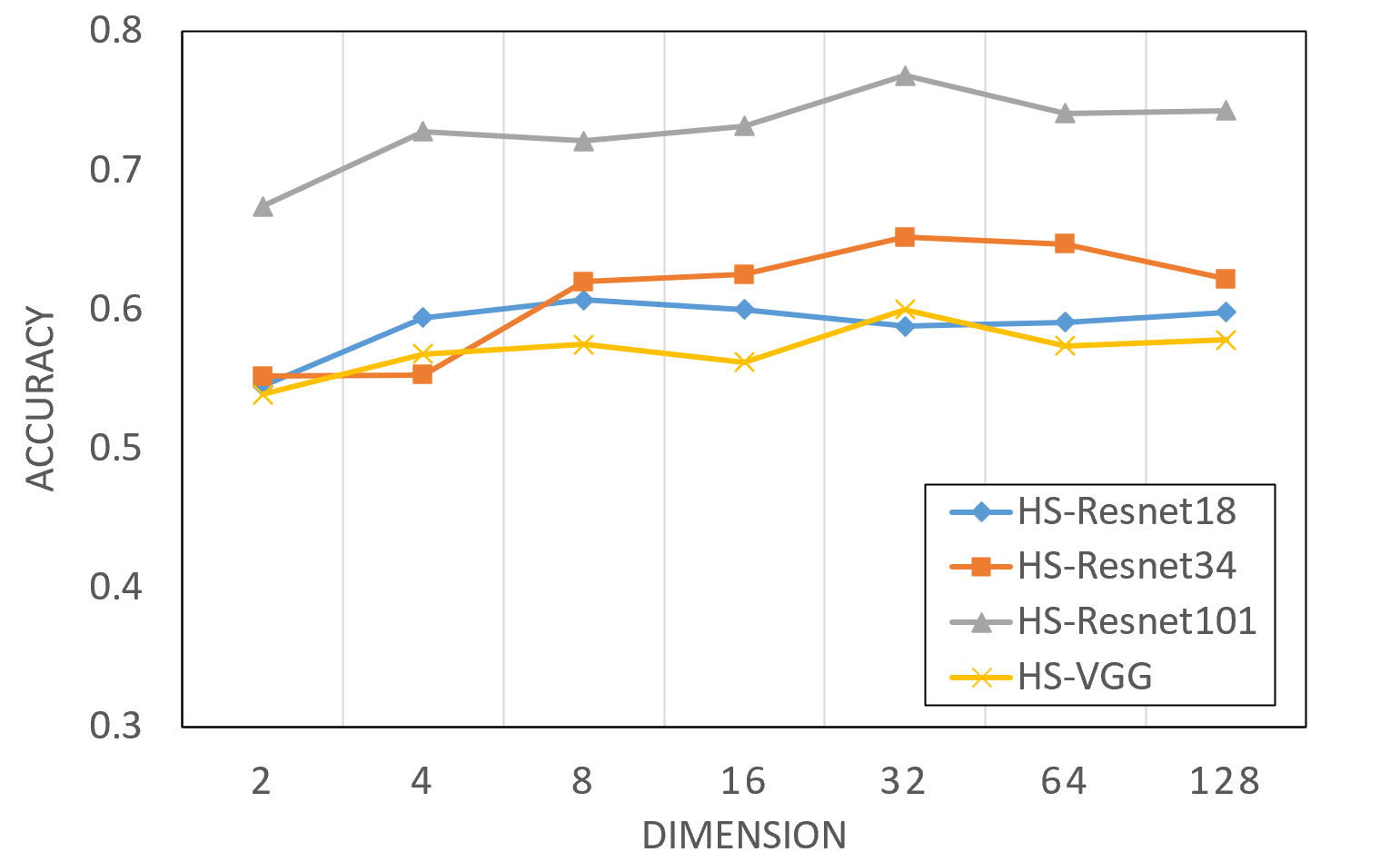}}
\vspace{-10pt}
\caption{The impacts of the embedding dimensions used in HSNs. }
\label{fig:dimension}
\vspace{-10pt}
\end{figure}

\noindent {\textbf{Ball dimension}}
%As the volume of hyperbolic space increases exponenlly with the radius, the embedding dimension needed to represent the feature embeddings can be much lower than that in Euclidean space. To demonstrate this, we investigate the performance of our model regarding different semantic embedding dimensions. In Figure 4, we compare the performance of  with DeViSE and ConSE. 
%On the contrary, the performance DeViSE and ConSE, which learns the embeddings in Euclidean space, decreases to 0 as both models cannot converge in training with 10-dimensional semantic embeddings. The results clearly show the advantages of learning embeddings in hyperbolic space.
Due to the exponential growth of hyperbolic space volume with radius, the necessary embedding dimension for representing features can be notably lower than that required in Euclidean space. We investigate the performance of HSNs regarding different {\Poincare}    ball embedding dimensions, as shown in Figure \ref{fig:dimension}. 
From Figure \ref{fig:dimension},  it can be seen that there is an improvement trend in performance as the dimension of the image embedding increases from 2 to 32.  However, with a further increase, the performance begins to decline. This is likely because while initial gains from higher dimensions improve data complexity capture, excessive dimensions eventually cause overfitting, diminishing the model's effectiveness.

%As illustrated in Figure \ref{fig:dimension}, there is an improvement trend in performance when the dimension of the image embedding reaches 32. However, as the dimension increases to 64, there is a decline in performance.
%our model still achieves a satisfactory

\begin{table}[t]
	\centering
	\caption{Ablation study of each module.} 
\vspace{-10pt}
\scalebox{0.90}{
	\begin{tabular}{|c |c| c|  } 
		\hline\
		Method & ACCURACY & F1  \\ \hline
	    Siamese network&82.09  &  0.5947\\ \hline
        Siamese network+HL& 85.35 &0.7164 \\ \hline
        Siamese network+HL+FCLP& \textbf{86.83} & \textbf{0.7372} \\ \hline
	\end{tabular}}
 \label{table_ablation}
 \vspace{-15pt}
\end{table}

\noindent {\textbf{Hyperbolic module.}}
In this section, we conduct an ablation study to evaluate the contribution of each component in HSNs.  As shown in Table \ref{table_ablation}, the integration of Hierarchical Learning (HL) into the baseline model leads to a significant improvement in its performance by 20\% in terms of F1-score. Moreover, the incorporation of Feature Clipping (FCLP) boosts the F1-score, from 0.7164 to 0.7372. These  improvements verify the effectiveness HL and FCLP.

\subsection{Benchmark evaluations}

%In this section, we first make comparisons with handcrafted features and deep features on CASME II, CASME, and SAMM databases, respectively. Then we conduct ablation analysis of the proposed SCA network. We finally evaluate the computational and storage cost.

In this section, we provide the baselines of TCD, and verify the effectiveness of HSN. Due to limited research on fine-grained change detection, the comparison methods are also limited.
The state-of-the-art (SoTA) method cost-effective tree-blossom recognition \cite{lee2023cost},  ResNet101, VGG16, InceptionV3 \cite{szegedy2015going}, and MobileNet \cite{sinha2019thin} are utilized. The results are shown in Table \ref{table_baseline}.
It can be seen the HSN outperforms all other listed methods. The HSN achieves the highest accuracy of $86.83\%$ and an F1-score of $0.7372$, significantly surpassing other approaches such as ResNet101, InceptionV3, MobileNet, and Tree-blossom recognition, by $24.0\%$, $29.9\%$, $60.5\%$, and $35.2\%$ in terms of F1-score, respectively.  These results highlight the superiority of the HSN in terms of precision and reliability in TCD. We can draw a conclusion that there is hyperbolicity in TCD and the HSN can better represent TCs with hyperbolicity. For more comparison results, please refer to the supplementary material.

\begin{table*}[htbp]
\centering
\caption{ Evaluation of CD-FAS among CASIA (C), Idiap Replay (I), MSU-MFSD (M), and Oulu-NPU (O) databases. }
\vspace{-8pt}
\scalebox{0.80}{
\begin{tabular}{|c|c|c|c|c|}
\hline
Method (\%) &\makecell[c]{ OCI $\rightarrow$ M   \\  (HTER$\downarrow$/AUC$\uparrow$)}    & \makecell[c]{ OMI $\rightarrow$ C \\ (HTER$\downarrow$/AUC$\uparrow$)} & \makecell[c]{ OCM $\rightarrow$ I \\ (HTER$\downarrow$/AUC$\uparrow$)} & \makecell[c]{  ICM $\rightarrow$ O \\ (HTER$\downarrow$/AUC$\uparrow$)} \\
\hline
SSDG-R \cite{jia2020single} & 14.65 / 91.93 & 28.76 / 80.91  & 22.84 / 78.67 & 15.83 / 92.13  \\
\hline
SSAN-R \cite{wang2022domain} & 21.79 / 84.06  & 26.44 / 78.84  & 35.39 / 70.13  & 25.72 / 79.37 \\
\hline
PatchNet \cite{wang2022patchnet} & 25.92 / 83.43  & 36.26 / 71.38  & 29.75 / 80.53 & 23.49 / 84.62 \\
\hline
SA-FAS \cite{sun2023rethinking} & {13.17}/ {94.17} & {24.48} / {85.55}  & {19.79} / {87.95}  & {14.23} / {93.29}  \\
\hline
 Ours & 
 \textbf{11.98}/ \textbf{94.33}  & \textbf{17.46} / \textbf{90.89}  & \textbf{18.75} / \textbf{89.54}  & \textbf{13.45} / \textbf{94.34} \\
\hline
\end{tabular}}
\label{tab:evaluation_convergence}
\vspace{-10pt}
\end{table*}

\begin{figure}
\centerline{\includegraphics[width=8.0cm]{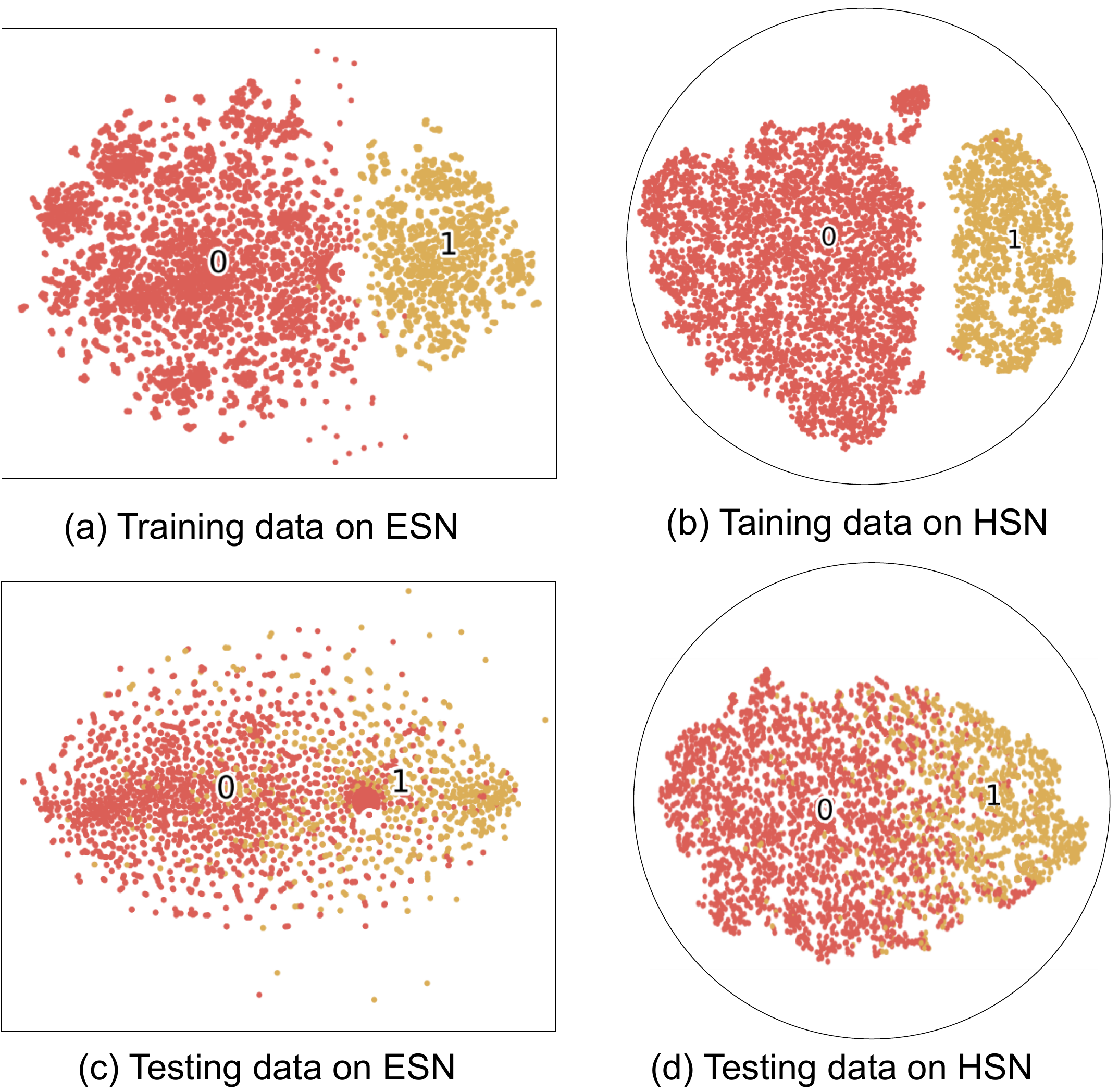}}
\vspace{-5pt}
\caption{The visualizations on Euclidean and hyperbolic spaces with t-sne \cite{van2008visualizing}, respectively. ESN and HSN represent Siamese networks in Euclidean space and hyperbolic space, respectively. }
\label{fig:vis}
\vspace{-20pt}
\end{figure}

\begin{table}[t]
	\centering
\vspace{-5pt}
	\caption{Comparisons in terms of ACCURACY and F1-scores on  UAVTC.} 
 \vspace{-6pt}
  \scalebox{0.80}{
	\begin{tabular}{|c |c| c|  } 
		\hline\
		Method & ACCURACY & F1  \\ \hline
          % ResNet18 \cite{he2016deep}&78.40  &  0.5331\\ \hline
          %  ResNet34 \cite{he2016deep}& 77.15 &0.5165  \\ \hline
            ResNet101 \cite{he2016deep}& 82.09 & 0.5947 \\ \hline 
            VGG16 \cite{simonyan2014very}& 79.21 &  0.5716        \\ \hline
            InceptionV3 \cite{szegedy2015going}&78.48&  0.5673        \\ \hline
            MobileNet \cite{sinha2019thin}& 76.83 &  0.4592    \\ \hline
            Tree-blossom recognition \cite{lee2023cost}& 78.82 &  0.5413       \\ \hline
            HSN (Ours)&  \textbf{86.83}&\textbf{0.7372}\\ \hline
	\end{tabular}}
 \label{table_baseline}
 \vspace{-10pt}
\end{table}

\begin{figure}
\centerline{\includegraphics[width=7.5cm]{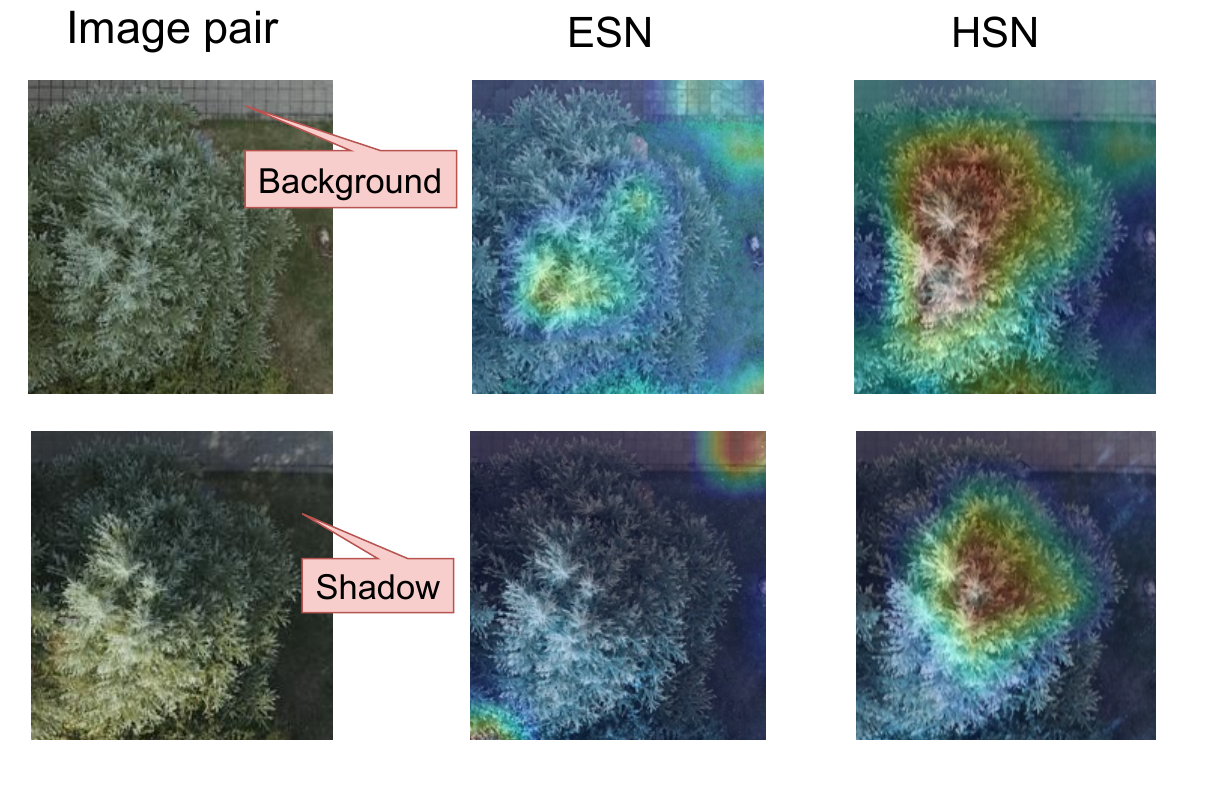}}
\vspace{-10pt}
\caption{The visualizations of tree change with Gradcam. }
\vspace{-15pt}
\label{fig:viscam}
\end{figure}

Figure \ref{fig:vis} shows the visualizations of training and testing image embeddings utilizing t-SNE for dimensionality reduction and visualization \cite{van2008visualizing}. The image embeddings were extracted by the Siamese network in Euclidean space and hyperbolic space with Resnet101 as the backbone. From Figure \ref{fig:vis}, it is evident that the HSN exhibits tighter clustering of same-category data points compared to the Euclidean-based networks, indicating that the network is learning more uniform and distinctive features in hyperbolic space. Furthermore, Figure \ref{fig:viscam} displays class activation maps of TC \cite{selvaraju2017grad}. The HSN appears to effectively concentrate on the tree regions, successfully avoiding misinterpretations caused by shadow and background. More visualization examples are shown in the supplementary material. 

\subsection{Complexity of Hyperbolic Operations}
Hyperbolic operations are  more computationally intensive compared to Euclidean ones. To mitigate this, we integrated hyperbolic operations with CNN, employing them primarily in the final layer to maintain efficiency. On average, one iteration takes 0.4358 seconds for the Euclidean siamese network and 0.4437 seconds for HSN. It indicates while hyperbolic operations may incur some overhead, our approach balances computational complexity with performance gains.

%applying hyperbolic spaces reveals a concentration of all embeddings within a confined region of the entire {\Poincare} ball. 

%\subsection{\textcolor{black}{Contribution and future work}}

%The contribution of this research lies in the establishment of a UAVTC for monitoring fine-grained TCs over time and the design of HSN to effectively detect and represent these changes in hyperbolic space, addressing issues related to image quality fluctuations due to weather and the detailed hierarchical nature of TCs.

%

%HSN learns robust embeddings in hyperbolic space, effectively capturing the intricate hierarchical structures of CD-FAS challenges and representing complex variations and attack types  in a compact low-dimensional space.

\subsection{Evaluation on CD-FAS}
To further prove the effectiveness and impacts of HSN,  we evaluate its performance on CD-FAS tasks.
CD-FAS differentiates genuine human faces from counterfeit representations in biometric authentication systems across different domains. CD-FAS involves identifying complex variations and changes amoung extrinsic factors, such as face coverings and lighting conditions, as well as intrinsic factors such as skin texture and material properties. Four popular benchmark datasets are utilized: Oulu-NPU (O) \cite{boulkenafet2017oulu}, CASIA (C) \cite{zhang2012face}, Idiap Replay Attack (I) \cite{chingovska2012effectiveness}, and MSU-MFSD (M) \cite{wen2015face}, highlighting its significance in AI. Leave-one-out test protocol to evaluate their cross-domain generalization.

 %Following prior works, we treat each dataset as one domain and apply the leave-one-out test protocol to evaluate their cross-domain generalization. Specifically, we refer to OCI$\rightarrow$M as the protocol that trains on Oulu-NPU, CASIA, Idiap Replay attack and tests on MSU-MFSD. OMI$\rightarrow$C, OCM$\rightarrow$I and ICM$\rightarrow$O are defined in a similar fashion.

For fair comparisons with SoTA methods \cite{jia2020single, wang2022domain, wang2022patchnet, sun2023rethinking}, we use the same ResNet-18 backbone. We adapt the learning strategy and evaluation protocol of SA-FAS, employing metrics such as Half Total Error Rate (HTER) and Area Under the Curve (AUC). More implementation details are provided in the Supplementary materials.

From Table \ref{tab:evaluation_convergence}, we can see that our method based on HSN outperforms SoTA methods in CD-FAS \cite{jia2020single, wang2022domain, wang2022patchnet, sun2023rethinking} in terms of HTER and AUC. The results
 further validate the superiority of HSN on change identification, demonstrating its broad impact on the AI community. Additional results and analysis are included in Supplementary materials.

%The input images are cropped using MTCNN \cite{zhang2016joint} and resized to $256\times256$. For fair comparisons with state-of-the-art (SoTA) methods \cite{jia2020single, wang2022domain, wang2022patchnet, sun2023rethinking}, we use the same ResNet-18 backbone. We train the network with the SGD optimizer and an initial learning rate of $5\times10^{-3}$, which is decayed by 2 at epoch 40 and 80, and the total training epoch is 100 in most set-ups\footnote{Detailed settings may vary based on experimental protocols.}. We set the weight decay as $5\times10^{-4}$ and the batch size as 96 for each training domain.

%\subsubsection{Evaluation Metrics}
%We evaluate the model performance using three standard metrics: Half Total Error Rate (HTER), Area Under Curve (AUC), and True Positive Rate (TPR95) at a False Positive Rate (FPR) of 5\%. While HTER and AUC assess the theoretical performance, TPR at a certain FPR is adept at reflecting how well the model performs in practice.

%\vspace{2cm} 

\vspace{-5pt}
\section{Conclusion}
\vspace{-5pt}
This paper presents a UAV-camera Tree change (UAVTC) dataset for fine-grained monitoring trees over time and a novel Hyperbolic Siamese Network (HSN) for accurately detecting tree changes. The HSN is able to effectively distinguish physiological tree changes from  extrinsic changes driven by  environment variants. Our extensive experiments confirm the hierarchical nature of tree change and  the superior performance of the HSN illustrates a significant advancement in the field of precision forestry. Our multidisciplinary research on tree monitoring sets new benchmarks and introduces cutting-edge AI technologies, enhancing ecosystem and biological understanding. Furthermore, the HSN generalizes well to  change-identification applications, such as cross-domain face anti-spoofing tasks,  underscoring its significance in AI.

In the future, we aim to enhance the dataset to encompass tree segmentation and classification tasks. This expansion could significantly benefit biodiversity research and lead to more precise and all-encompassing approaches in forest management.

%Moreover, the UAVTC dataset is tailored for detecting tree changes. 

%\input{supp}

{
    \small
    \bibliographystyle{ieeenat_fullname}
    \bibliography{main}
}

%\input{sec/4_suppl}
\end{document}